\documentclass{bmvc2k}
\usepackage[misc]{ifsym}
\usepackage{graphicx}
\usepackage{subcaption}
\usepackage{booktabs}
\usepackage{caption}
\usepackage{amsmath} 
\usepackage{amssymb} 
\usepackage{hyperref}
\usepackage{bm}

\title{ControlEdit: A MultiModal Local Clothing Image Editing Method}

\addauthor{Di Cheng}{202220001093@stu.bift.edu.cn}{1}
\addauthor{Yingjing Shi$^{(\textrm{\Letter})}$}{20150015@bift.edu.cn}{1}
\addauthor{Shixin Sun}{202220001094@stu.bift.edu.cn}{1}
\addauthor{Jiafu Zhang}{202320001090@stu.bift.edu.cn}{1}
\addauthor{Weijing Wang}{20230001092@stu.bift.edu.cn}{1}
\addauthor{Yu Liu}{202320001091@stu.bift.edu.cn}{1}
\addinstitution{
School of Arts {\&} Sciences,\\
 Beijing Institute of Fashion\\
 Technology, Beijing, China
}

\runninghead{Student, Prof, Collaborator}{BMVC Author Guidelines}

\begin{document}

\maketitle

\begin{abstract}
Multimodal clothing image editing refers to the precise adjustment and modification
of clothing images using data such as textual descriptions and visual images as control conditions, which effectively improves the work efficiency of designers and reduces
the threshold for user design. In this paper, we propose a new image editing method ControlEdit, which transfers clothing image editing to multimodal-guided local inpainting of clothing images. We address the difficulty of collecting real image datasets by
leveraging the self-supervised learning approach. Based on this learning approach, we
extend the channels of the feature extraction network to ensure consistent clothing image
style before and after editing, and we design an inverse latent loss function to achieve
soft control over the content of non-edited areas. In addition, we adopt Blended Latent
Diffusion as the sampling method to make the editing boundaries transition naturally
and enforce consistency of non-edited area content. Extensive experiments demonstrate
that ControlEdit surpasses baseline algorithms in both qualitative and quantitative evaluations.The code and pretrained models will be available on GitHub. Check \url{https://github.com/cd123-cd/ControlEdit}

\end{abstract}

\section{Introduction}
\label{sec:intro}

Clothing image editing refers to the process of users making simple modifications to a given clothing image and obtaining a realistic modified physical image through algorithms. The clothing image editing model allows designers to interactively translate design concepts into real images. At the same time, ordinary users are allowed to communicate with professional clothing designers to optimize the personalized clothing customization process using physical images as a reference. The majority of previous clothing image editing methods are based on GAN-based generative approaches \cite{zhu2017your,hsiao2019fashion++,men2020controllable,lin2023fashiontex,dong2020fashion,ak2019attribute}. Recently, the FICE method\cite{pernuvs2023fice} has combined GAN with the CLIP model to achieve semantic constraints, thereby achieving fine-grained content editing. Despite significant progress, training a new usable model is extremely dependent on dataset. Furthermore, the model's generation capability is restricted due to the scarcity of attribute text annotations. In addition, previous work has focused on attribute-guided image editing\cite{zhu2017your,hsiao2019fashion++,men2020controllable,pernuvs2023fice,lin2023fashiontex,ak2019attribute} and sketch-guided image\cite{dong2020fashion} editing. Although attribute-guided image editing can convey specific attributes of clothing such as style, color, and pattern, it may not provide enough geometric information to generate images consistent with what the user has in mind; sketch-guided image editing can assist in presenting the shape and layout of images, however, it is difficult to control semantic information such as image style. Hence, it is crucial to develop a multimodal clothing image editing method that combines sketches, text, and real images to enhance the clothing editing process.

Recently, large-scale language image (LLI) models\cite{zhang2021ernie,feng2023ernie,wu2022nuwa,ramesh2022hierarchical,ding2021cogview,kang2023scaling,rombach2022high} have shown exceptional generation capabilities. These models allow for fine-tuning using various methods to adapt to downstream tasks in multiple domains\cite{zhang2023adding}. There is relatively little work on multimodal clothing images editing in the fashion field, MGD\cite{baldrati2023multimodal} and RBI\cite{kim2023reference} have successfully fine-tuned pre-trained models to generate high-quality clothing images, making clothing image editing tasks possible. However, the commonly used LLI model introduces some randomness in the image generation process, which can result in slight differences in each generated image. The occurrence of "the slightest nudge causes the widest chain reaction" has a comprehensive impact on the final generated image. Therefore, it is particularly important to develop a controllable multimodal image editing method to optimize clothing design.
\begin{figure}[h]
  \centering
\captionsetup{skip=5pt}
  \includegraphics[width=0.9\textwidth]{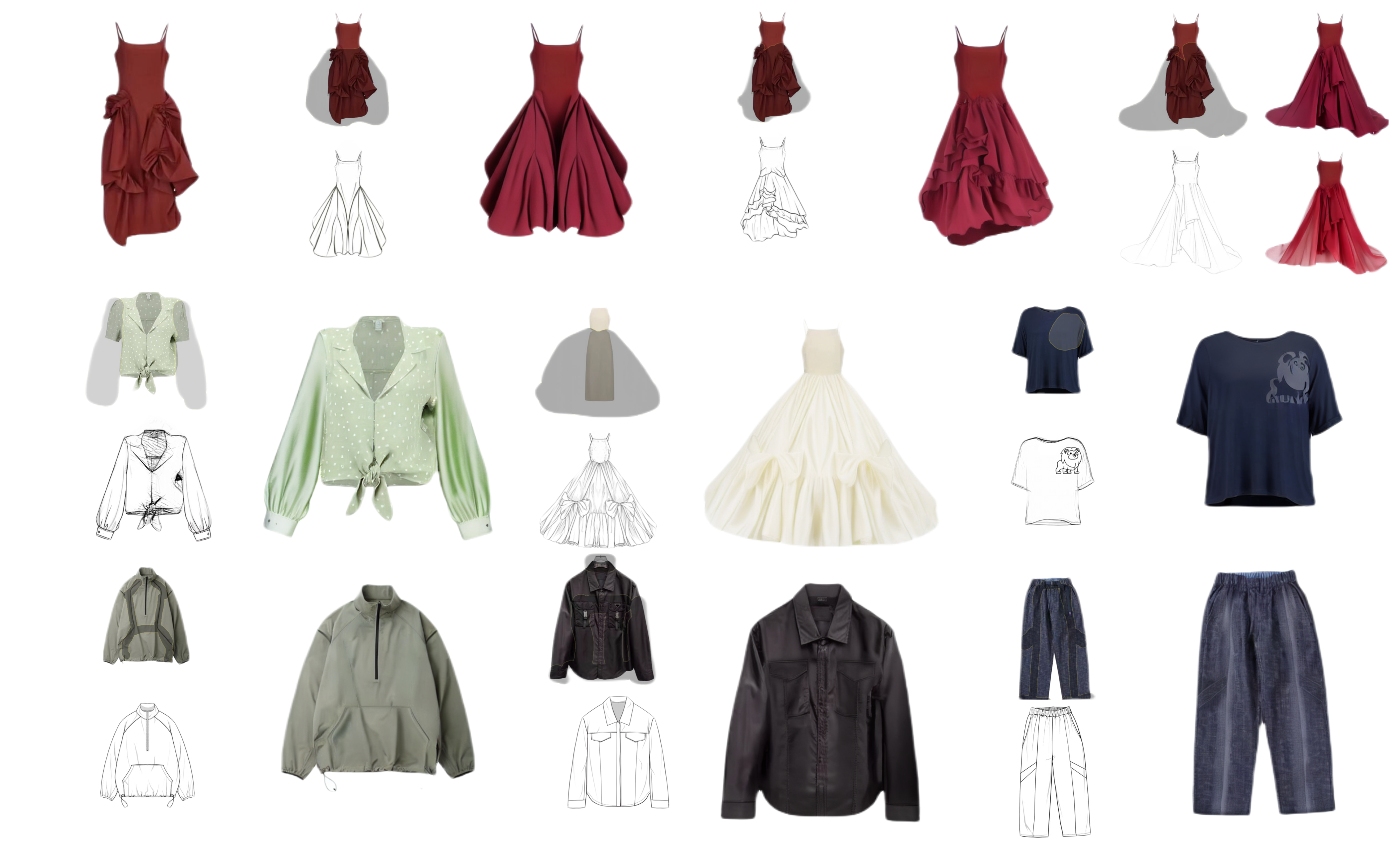}
  \caption{ControlEdit. Users can edit clothing by drawing conditional images.The first two rows of images are edited by regular users, while the last row is modified by professional fashion designers.}
  \label{fig:image 1}
\end{figure}        

In this paper,we propose a local editing method based on sketches and text and define a new task for multimodal conditioned fashion image editing. This method allows us to guide the generative process via multimodal prompts, maintaining the controllable, realistic, and plausible nature of the edited images(Fig.\ref{fig:image 1}).  The key challenges of multimodal clothing image edit are: (1) It is difficult to collect sufficient paired real clothing images before and after modification for training. (2) The utilization of masks serves as the primary measure to maintain the integrity of non-edited regions, however, it introduces artifacts in the transition areas. Balancing the preservation of unchanged regions of clothing images alongside ensuring natural transitions between edited and non-edited regions while maintaining stylistic consistency poses a significant challenge. (3) Unlike image translation tasks, our task not only conducts domain translation from sketch to physical image, but also requires further reasonable fusion between the generated physical image and the source physical image.

The main contributions of this paper are as follows: (1) 
We propose ControlEdit, the first multimodal local editing method for clothing images, which is based on Controlnet. ControlEdit leverages sketches, natural language, and masked source images to guide image generation. Our editing process aligns with the typical practices of designers when making modifications.(2) We propose an inverse latent loss function, which optimizes the native Controlnet loss function and promotes consistency in non-edited area content. (3) We perform a mask fusion operation on the generated features and the source image features at each inference step in the latent space, avoiding issues such as unnatural pixel space mask transitions and inconsistent styles. (4) The above work has shown better image generation quality than the baseline model in benchmark testing.

\section{Related works}
{\bf GAN-based Clothe Image Edit} In order to generate real clothing images, existing methods based on generative adversarial networks\cite{creswell2018generative} usually map clothing control conditions to latent spaces and then perform clothing editing. Fashion++ \cite{hsiao2019fashion++}associates semantic segmentation maps with texture features and shape features. ADGAN\cite{men2020controllable} maps human attributes to latent space as independent code, and achieves attribute control through mixing and interpolation operations. FE-GAN\cite{dong2020fashion} and FashionGAN\cite{zhu2017your} encode control images into the synthesized parsing map, which guide the generation of clothing image details. FashionTex\cite{lin2023fashiontex} maps portrait, text and texture to latent space to obtain different latent vectors for manipulating image generation. FICE\cite{pernuvs2023fice} utilizes pre-trained GAN \cite{creswell2018generative}generators and CLIP models to implement semantic constraints. However, existing methods may encounter issues such as clothing image artifacts and lack of realism. This paper proposes an effective approach to address these issues by leveraging the robust generative capabilities of pre-trained models.

{\bf Diffusion-based Clothe Image Edit} The rapid development of diffusion models\cite{ho2020denoising} has been proven to surpass GANs, however, there is currently limited work on clothing image editing based on diffusion models. Text2Human\cite{jiang2022text2human} adds diverse text guidance to generate realistic texture portrait images based on human text for human body analysis. MGD\cite{baldrati2023multimodal} and \cite{kim2023reference} fine-tune pre-trained diffusion models to use reference images to complete missing areas while maintaining control condition guidance. DiffFashion\cite{cao2023difffashion} guides the denoising process through automatically generated semantic masks and pre-trained visual transformers (ViT)\cite{amir2021deep}, allowing for appearance transfer while preserving structural information. Our approach differs from the aforementioned method, which emphasizes the use of textual descriptions and sketches as conditions for virtual try-on tasks. Instead, we focus on directly editing the garments themselves.
\begin{figure}[h]
  \centering
  \includegraphics[width=1\textwidth]{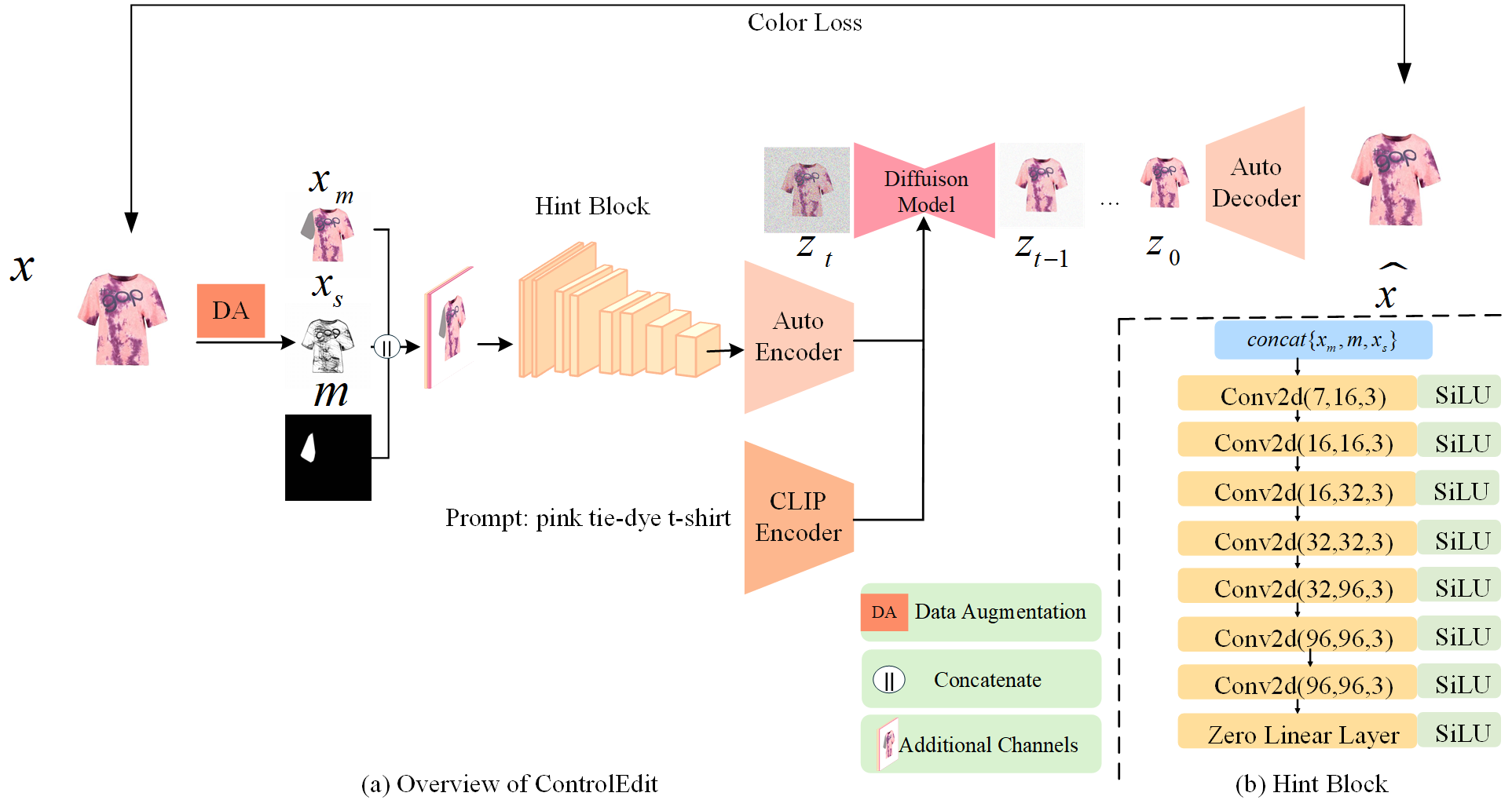}
  \captionsetup{skip=4pt}
  \caption{ControlEdit Network Architecture.}
  \label{fig:image 2}
\end{figure}
\section{Method}

During the process of local clothing image editing, given the image $x\in\mathbb{R}^{H\times W\times3}$ to be edited, the user adds the mask $m\in\{0,1\}^{\mathrm{H\times W}}$ on the image to simulate the modified area, where the value of 0 specifies the editable position, and the value of 1 ensures maximum consistent with $x$. The sketches drawn by users at masked positions are combined with the sketches extracted from the source image using Controlnet to obtain $x_{s}$ through mask fusion operations, $x_{m}$ representing the pre-editing state image. The training data is $\{(x_{s},x_{m},m,text),x\}$. Our aim is to conduct multimodal-based local image editing, wherein reference conditions are automatically integrated into the source image, ensuring that the resulting image appears controllable, realistic and plausible.

\subsection{Preliminary}
{\bf Controlnet} Our ControlEdit is an extension of Controlnet\cite{zhang2023adding}, which is the fine-tuning method that copies the weights of LDM\cite{rombach2022high} to "trainable copy" and "locked copy".
the locked copy retains the network capabilities learned from billions of images, while the trainable copy is trained on task specific datasets to learn conditional control, connected through zero convolution. Forward process: The extracted feature maps are fed into the autoencoder and converted into latent variable. Given the variance $\beta$, noise image $z_0$ is added until $z_T \sim N (0,1)$. The forward process is defined as follows:
\begin{equation}
q\left(z_t \mid z_{t-1}\right)= N \left(z_t ; \sqrt{1-\beta_t} z_{t-1}, \beta_t I \right)
\label{eq:formula 1}
\end{equation}

Reverse process: The reverse process can gradually remove noise by running reverse learning until a new sample is generated. Eq.\ref{eq:formula 2} allows for generating different reverse samples by changing the variance of the noise. Among them $\mu_\theta\left(z_t, t\right)$, $\Sigma_\theta\left(z_t, t\right)$ is the parameter for predicting Gaussian distribution.
\begin{equation}
p_\theta\left(z_{t-1} \mid z_t\right)= N \left(z_{t-1} ; \mu_\theta\left(z_t, t\right), \Sigma_\theta\left(z_t, t\right)\right)
\label{eq:formula 2}
\end{equation}

The loss function of Controlnet is shown in Eq.\ref{eq:formula 3}, where text prompts $c_t$ and $c_f$ are conditional feature maps, and $\epsilon_\theta(\cdot)$ is the denoising network.
\begin{equation}
\left. L _{c l d m}= E _{z_0, t, c_t, c_t, \epsilon \sim N (0,1)}\left[\| \epsilon-\epsilon_\theta\left(z_t, t, c_t, c_f\right)\right) \|_2^2\right]
\label{eq:formula 3}
\end{equation}

\subsection{ControlEdit}

The overall structure of ControlEdit is shown in Fig.\ref{fig:image 2}(a). Our task aims to generate target clothing images based on sketches, text, masks, and masked source images. We adopt Controlnet for initialization as the image prior to preserve the model's original controllability. Lacking a series of pre-edited real images, modified sketch images, and post-edited real images, we provide masked source images to the network to simulate pre-edited clothing images. The purpose is to allow the network to retain content in non-edited areas and provide color references for the generated areas when generating editing results; at the same time, in order to enhance the model's perception of editing positions, mask information is introduced to enable the network to better understand the spatial information of the target area of editing operations. Therefore, before the conditional features enter the denoising network, we extend the initial convolutional layer channel dimension of the conditional feature extraction network from 3 to 7 (i.e. 3+1+3), where $x_m \in R ^{\{H, W, 3\}}$ occupies three channels, $m \in\{0,1\}^{H \times W \times 1}$ occupies one channel, and $x_s \in R ^{\{H, W ; 3\}}$ occupies three channels. Providing more parameter space for the network enables it to more fully express the complex relationship between input conditions and output images, improving the flexibility and expressive power of the model.
\begin{figure}[h]
  \centering
  \includegraphics[width=0.4\textwidth]{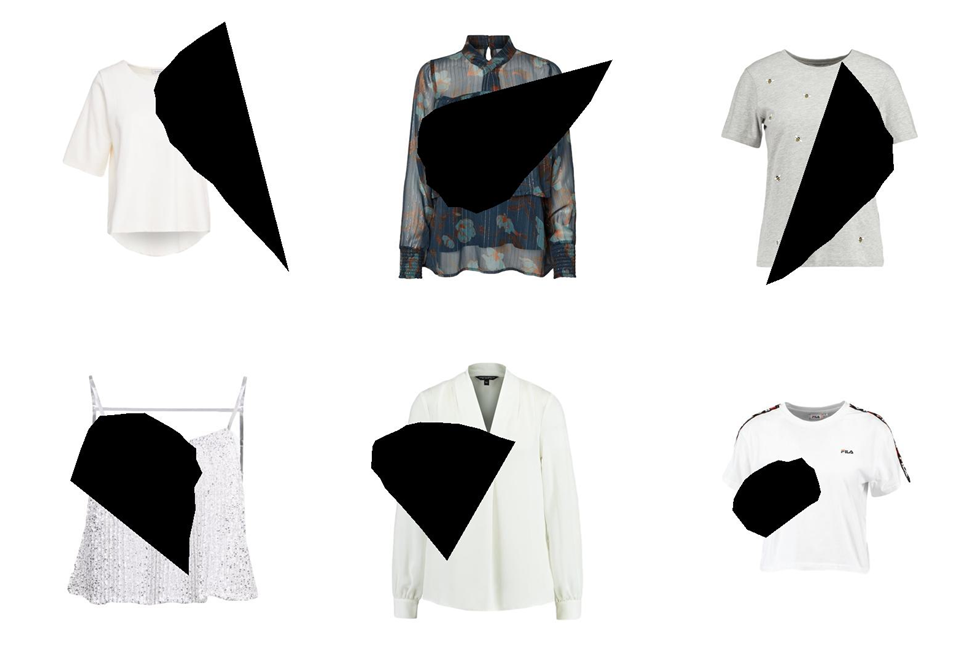}
  \caption{Masked Image Example.}
  \label{fig:image 3}
\end{figure}

{\bf Data augmentation} The shape and size of clothing image editing are subject to randomness, so adopting conventional-shaped masks results in the models being limited to learning simple mappings. Inspired by the Paint by Example\cite{yang2023paint}, we use the Bessel curve to sample 18 points and connect them to form a mask area of any shape, as shown in Fig.\ref{fig:image 3}. The generated mask area is closer to the actual editing operation, reducing the gap between training and testing, and enhancing the robustness.

{\bf Inverse latent loss function} On the one hand, ControlNet method based on sketches encounters certain limitations in color restoration and detail preservation during model training, as it lacks the RGB information in the non-edited regions. On the other hand, the encoder of ControlNet performs multiple downsampling operations, which further aggravates the information loss. To ensure that the generated image aligns with the ground truth and that the network structure has RGB information of the source image, we introduce the masked source images into the feature extraction network. These images provide RGB information for non-edited regions, while masks prevent the leakage of the content that the model needs to generate. The original Controlnet loss function is unable efficiently to bridge the gap between the editing and the non-editing domain. we propose the inverse latent loss function to force the editing model to pay more attention to maintaining the overall structure of the image and the consistency of the content in non-editing areas during the editing process. The image features predicted by the model are decoded into pixel space by the image decoder, and then we calculate L2 Euclidean distance between the decodeded image and the source image, which is as part of the total loss, as shown in Eq.\ref{eq:formula 5}.We modify $c_f$ in Eq.\ref{eq:formula 3} of $L_{c l d m}$ to be $c_f=E\left[\operatorname{cat}\left(x_m, x_s, m\right)\right]$, where $c_f$ represents the conditional feature map extracted by the encoder after concatenating the sketch, mask,and masked source image and $\hat{x}$ is the sampled image, which ensures the quality and realism of editing results.

\begin{equation}
L _{p i x}=\|x-\hat{x}\|^2
\label{eq:formula 4}
\end{equation}
\begin{equation}
L = L _{c l d m}+ L _{p i x}
\label{eq:formula 5}
\end{equation}
\begin{figure}[h]
  \centering
  \includegraphics[width=1\textwidth]{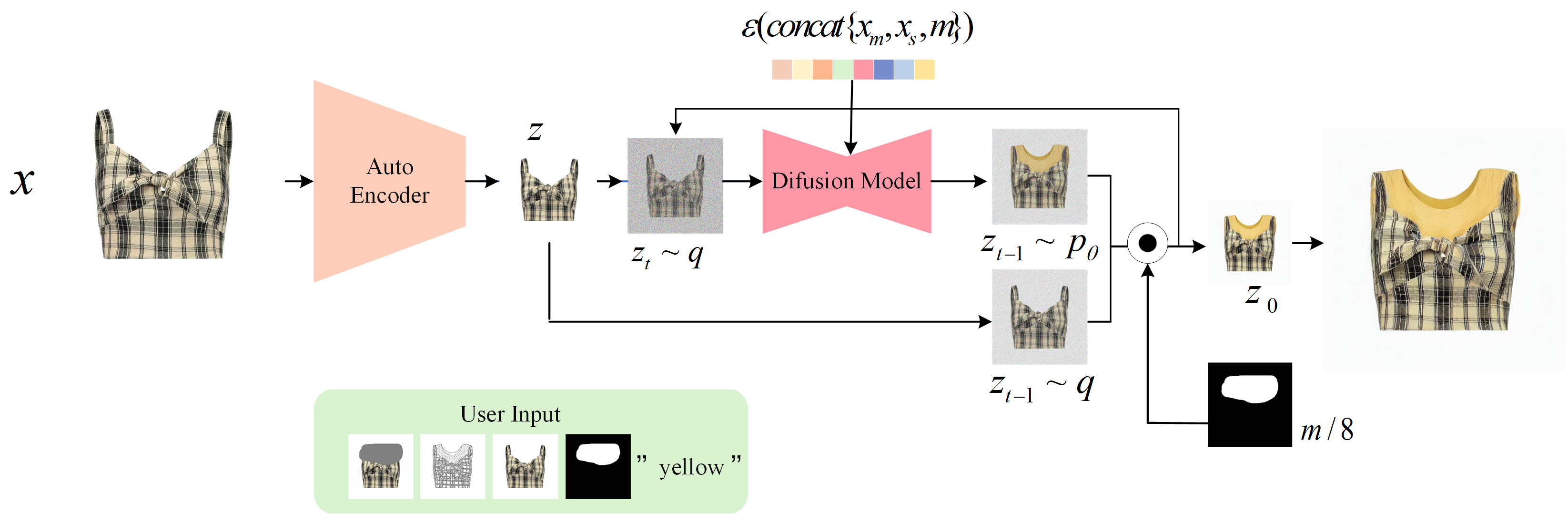}
  \captionsetup{skip=4pt}
  \caption{Image inference network structure.}
  \label{fig:image 4}
\end{figure}
{\bf Latent mask for sampling} To further ensure the natural transition, we utilized the Blended Latent Diffusion\cite{avrahami2023blended} sampling method in the inference stage. Through modifying latent variables in each denoising step and forcing the parts outside the mask to remain unchanged, it ensures that the colors of non-editing areas naturally transition to the editing area, maintaining the global color consistency. As shown in Fig.\ref{fig:image 4}, in the denoising step, we adopt the features of $text$ ,$x_m$, $x_s$, and $m$ as conditioning inputs for the Unet to obtain the latent variables of the editing area. For reverse steps, $z_{t-1}^{\text {old}}$ is sampled by the source image feature, and a certain level of noise is added to $z$ according to Eq.\ref{eq:formula 6}.
\begin{equation}
z_{t-1}^{\text {old }} \sim N \left(\sqrt{\bar{\alpha}_t} z_0,\left(1-\bar{\alpha}_t\right) I \right)
\label{eq:formula 6}
\end{equation}
Meanwhile, by using Eq.\ref{eq:formula 7} to process the noise of the original latent variable to the current noise level, we obtained the non-edited region latent variable with noise.
\begin{equation}
z_{t-1}^{\text {new }} \sim N \left(\mu_\theta\left(z_t, t\right), \Sigma_\theta\left(z_t, t\right)\right)
\label{eq:formula 7}
\end{equation}

Subsequently, we mixed these two latent variables by adjusting the mask size according to Eq.\ref{eq:formula 8}. It allows images to efficiently reduce obvious boundaries, making the overall effect more uniform.
\begin{equation}
z_{t-1}=m^{\prime} \odot z_{t-1}^{\text {old }}+\left(1-m^{\prime}\right) \odot z_{t-1}^{\text {new }}
\label{eq:formula 8}
\end{equation}

\section{Experiments}
\subsection{Evaluations}
{\bf Datasets} We adopt the text prompts and real clothing images from the MGD\cite{baldrati2023multimodal} dataset, and extract clothing sketches from real clothing images by Controlnet. The MGD multimodal dataset consists of 11647 pairs of clothing images. We select 9394 images as training data and 2000 images as subsequent model evaluation data. In order to truly test the editing and robustness of the model, we make varying degrees of modifications to these 2000 sketches, such as adding patterns, deleting sleeves, and changing round necks into suit collars etc.

{\bf Quantitative evaluation} In the task of image editing, the absence of precise evaluation metrics prompts the adoption of four metrics derived from the domain of generation for assessment purposes. FID\cite{heusel2017fid} measures the distribution similarity between the real image and the generated image by comparing the mean and variance of image features. LPIPS\cite{zhang2018LPIPS} is used to evaluate the perceptual difference between two images. Pre\_error\cite{zhu2023preerror} is the L2 distance between the non-edited regions of the generated image and the source image. The evaluation index for measuring the magnitude of image changes in non-edited areas. CLIP Score\cite{radford2021clipscore} is used to evaluate the semantic consistency between the generated image and the reference image.

\subsection{Results and analysis}
{\bf Baselines} To our knowledge, this is the first time that  diffusion models have been used for local image editing based on sketches and text. We choose three baseline models, and all methods are evaluated using the same 2000 images from the MGD dataset to sure fairness. 1) Controlnet. We use the edited sketch as a condition to represent the target content. 2)SD Inpainting. 3) Blended Latent Diffusion. 4)Uni-paint\cite{unipaint}. We use text prompts to represent the target content and masks to represent the generation area.
\begin{figure}[h]
  \centering
  \includegraphics[width=0.8\textwidth]{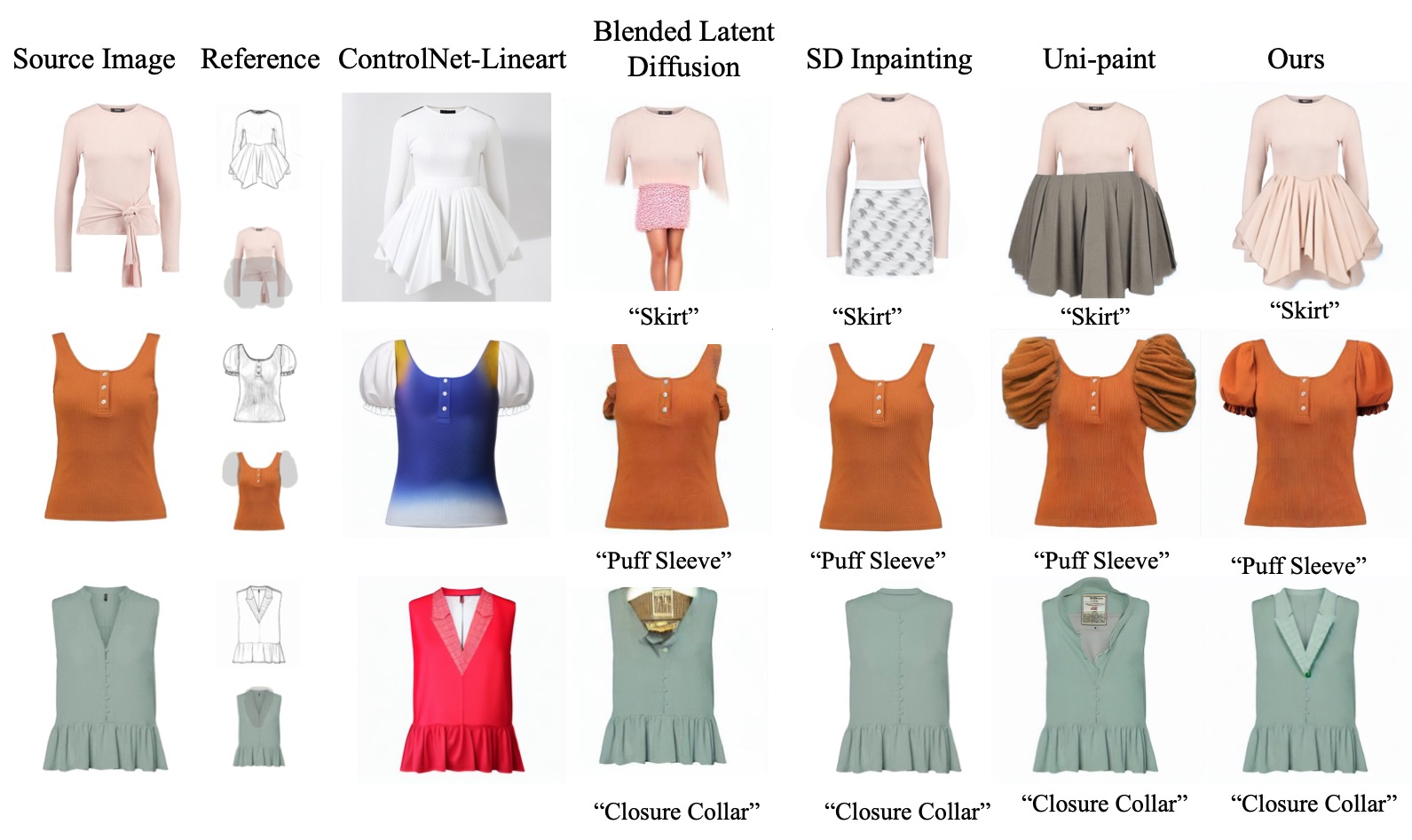}
  \captionsetup{skip=4pt}
  \caption{Qualitative comparison. SD Inpainting, Blended Latent Diffusion and Uni-paint synthesize clothing images driven by texts at the down side.}
  \label{fig:image 5}
\end{figure}
\begin{table}[h]
\centering
\captionsetup{skip=5pt}
\caption{Quantitative results of baseline model on 2000 images of size 512 x 512.}
\label{tab:animal-diet-size}
\small
\begin{tabular}{lcccc}
  \toprule
  Method & FID$\downarrow$ & LPIPS$\downarrow$ & Pre\_error$\downarrow$ & CLIP Score$\uparrow$  \\
  \midrule
  Controlnet-Lineart\cite{zhang2023adding} & 10.689 & 0.1219 &930.446 & 80.564  \\
  SD Inpainting \cite{rombach2022high}& 9.970 & 0.0938 & $\mathbf{1.322}$	& 78.318  \\
  Blended Latent Diffusion \cite{avrahami2023blended}& 7.49 & 0.1392 & 169.03 & 81.243  \\ 
  Uni-paint & 8.669 &0.0903&182.14& $\mathbf{81.994}$\\
  Ours & $\mathbf{4.569}$ & $\mathbf{0.0497}$ & 80.672 & 81.684  \\
  \bottomrule
\label{tab:table 1}
\end{tabular}
\end{table}

{\bf Qualitative analysis} We provide a qualitative comparison of these methods in Fig.\ref{fig:image 5}. Text-guided Blended Latent Diffusion, SD Inpainting and Uni-paint can generate images that match the description, however, it is difficult for regular language to specify fine-grained object appearances. Controlnet-Lineart focuses on translation between image domains with weak ability to maintain non-edited regions unchanged. Our method achieves unchanged non-editing areas, natural transitions, and generates areas that are loyal to the sketch and text conditions.

{\bf Quantitative analysis} Tab.\ref{tab:table 1} shows the quantitative comparison results. SD Inpainting achieves the lowest Pre\texttt{\_}error score, indicating its ability to preserve information from conditional images, however, it focuses on inpainting task and lacks guidance on conditional information for image generation. Our method exhibits superior performance on most metrics and has significant advantages over existing methods. Especially, our method has the lowest scores in both FID and LPIPS, indicating better fidelity and perceptual similarity between the generated image and the real image. In addition, the Pre\texttt{\_}error metric and CLIP-Score further confirm the effectiveness of our method in accurately reconstructing and preserving important features in generated images.
\begin{figure}[h]
  \centering
  \includegraphics[width=0.8\textwidth]{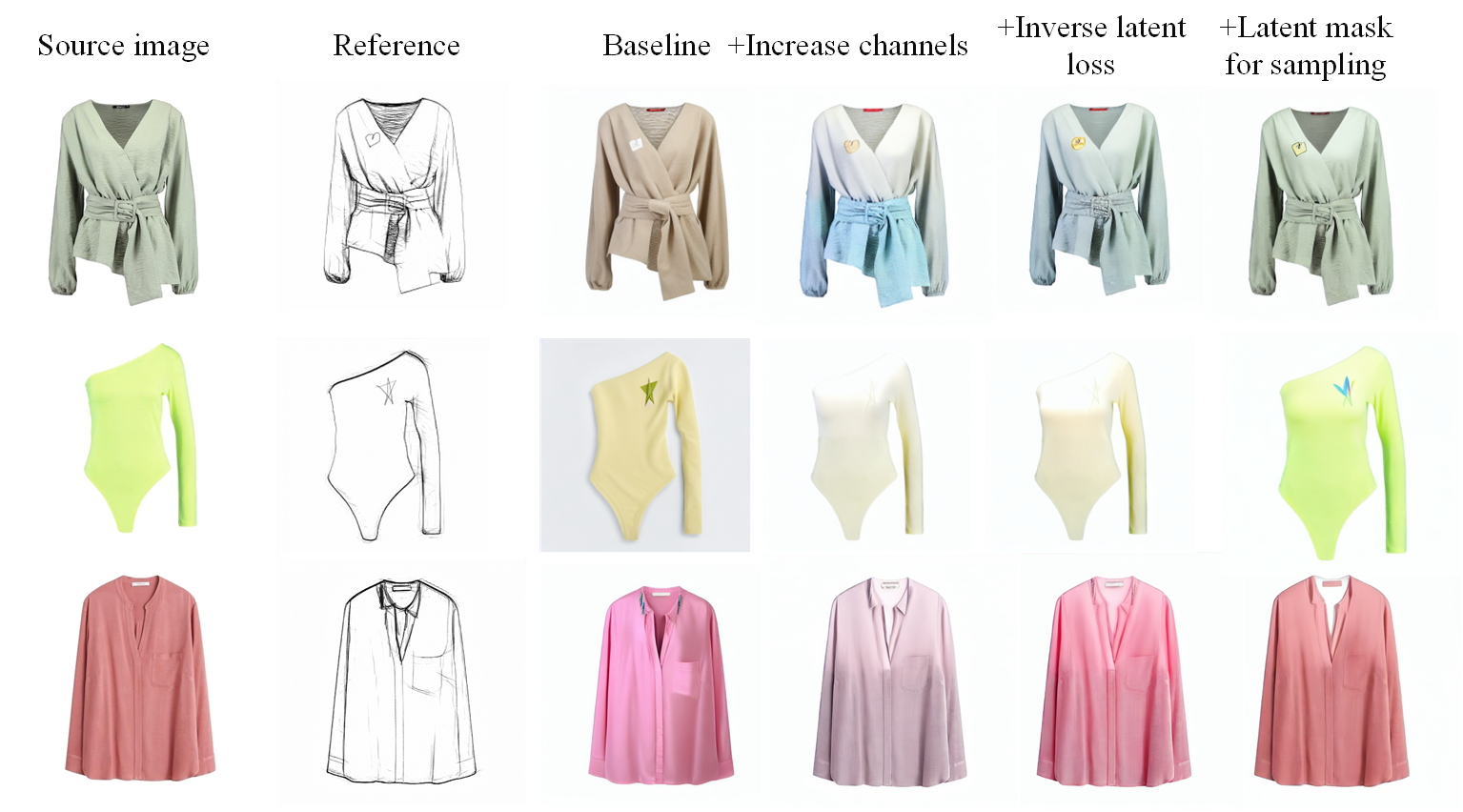}
  \caption{Visual ablation studies of individual components in our approach.}
  \label{fig:image 6}
\end{figure}

\begin{figure}[htbp]
  \centering
  \begin{minipage}[b]{0.4\linewidth} 
    \centering
    \includegraphics[width=\linewidth]{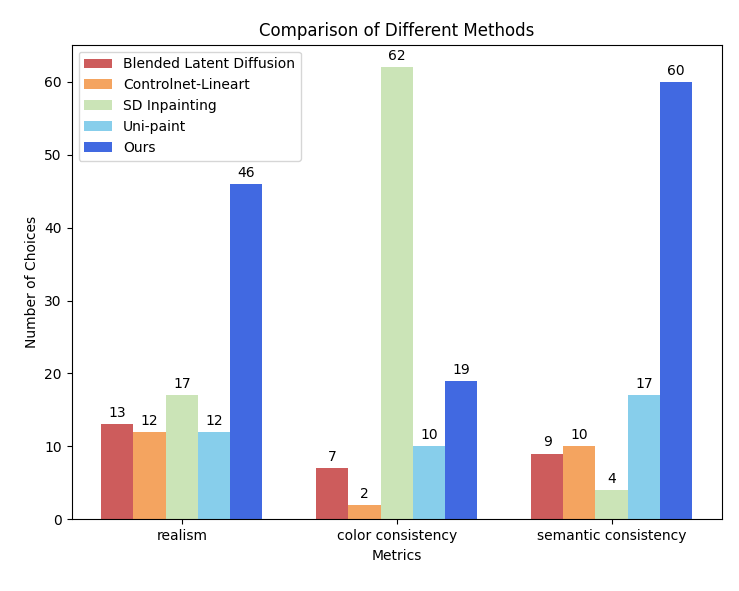}
    \label{fig:sub1}
  \end{minipage}
  \hspace{0.05\linewidth} 
  \begin{minipage}[b]{0.4\linewidth} 
    \centering
    \includegraphics[width=\linewidth]{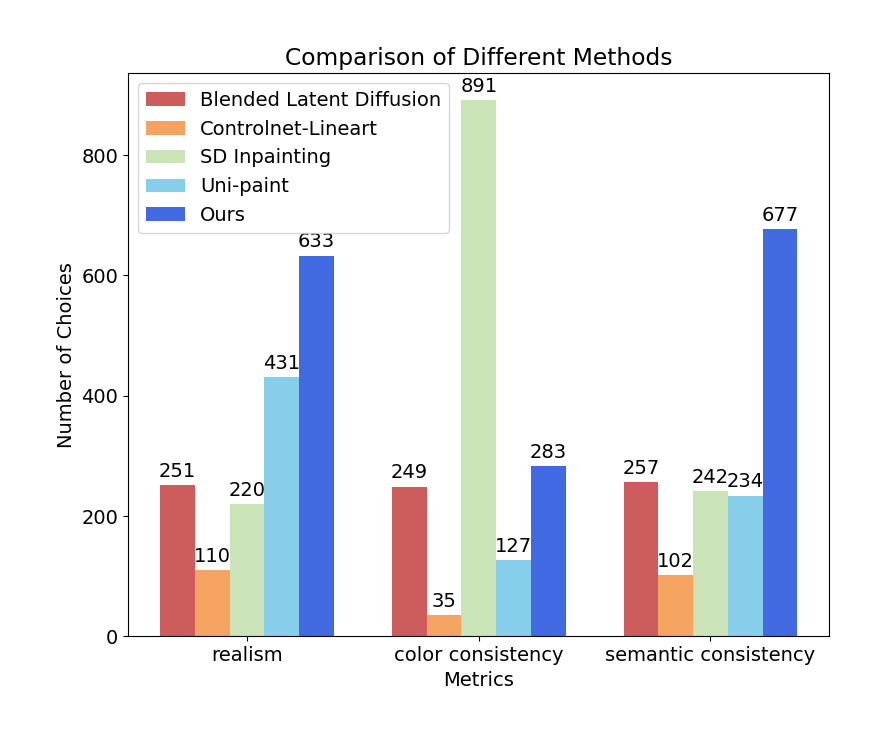}
    \captionsetup{width=0.8\linewidth} 
    \label{fig:sub2}
  \end{minipage}
  \caption{User study results. We compare our ControlEdit with three baselines.}
  \label{fig:main}
\end{figure}

\begin{table}[h]
\centering
\captionsetup{skip=5pt}
\caption{Quantitative results of ablation on 2000 images of size 512 x 512.}
\label{tab:table 1}
\small
\begin{tabular}{p{4cm}cccc}
  \toprule
  Method & FID$\downarrow$ & LPIPS$\downarrow$ & Pre\_error$\downarrow$ & CLIP Score$\uparrow$  \\
  \midrule
  Baseline & 10.689 & 0.1219 & 930.446 & 80.564  \\
  + Increase channels & 5.773 & 0.0739 & 292.1	& $\mathbf{81.704}$  \\
  + Inverse latent loss & 5.388 & 0.0684 & 231.16 & 81.693  \\
  + Latent mask for sampling & $\mathbf{4.569}$ & $\mathbf{0.0497}$ & $\mathbf{80.772}$ & 81.684  \\
  \bottomrule
\label{tab:table 2}
\end{tabular}
\end{table}

{\bf User study} For models trained on the MGD dataset, participants are presented with source images, baseline-generated images, and images generated by our model. They are asked to select images based on three criteria: (1) realism, (2) color consistency, and (3) semantic consistency.
We designed two approaches: 1) We select 10 participants to choose an image that best meets the aforementioned criteria. 2)We survey 100 participants, over half of whom are art majors, allowing them to select multiple images that meet the above criteria. As shown in Fig.\ref{fig:main}, whether single(left) or multiple(right) selections are allowed, our method exhibite superiority in human evaluation. Although the SD Inpainting method performe better in maintaining color consistency, it could not accurately generate semantic adjustments in clothing portions.
\subsection{Ablations} To achieve multimodal local editing, we introduced three methods, namely Increase channels, Inverse latent loss, and Latent mask for sampling. 1) We  represent Controlnet-Lineart as the baseline. 2) We have extended the Controlnet channels to express the complex relationship between input conditions and output images. 3) To ensure that the generated image matches the color of the source image, we add a new loss to the total loss. 4) To further ensure natural image transitions and color consistency, we use the Latent mask sampling method.
We present the results in Tab.\ref{tab:table 2} and Fig.\ref{fig:image 6}.The baseline solution generates images with a significant color difference between the generated image and the source image. By increasing channels, the image gradually approaches the source image in non-edited areas and transitions naturally, while the color style is inconsistent with the source image. When further adding loss, it alleviates the problem of non-editing area changes. Finally, using Latent mask further ensures that the color style of the generated image is consistent with that of the source image and the image transitions are natural, greatly improving the overall image quality and achieving the best performance.
\section{Conclusion} We propose ControlEdit, which is the first pre-trained diffusion model used for local image editing methods based on sketches and text. Our method overcomes the problem of insufficient collection of clothing datasets through self-supervision. Our proposed loss function effectively preserves the details of non-edited regions. Numerous experiments have clearly demonstrated that ControlEdit outperforms existing methods in many metrics, achieving high-quality and realistic results. We hope that this work can serve as a solid baseline and contribute to supporting future research in the field of clothing image editing.
\section*{Acknowledgement} This work is supported in part by the Open Research Project of Hubei Provincial Engineering Research Center for Intelligent Textile and Apparel (2023HBITF01), the Strategic Cooperation Agreement between Cloudsky Information Technology Co. and Beijing Institute of Fashion Technology (Project No. H2024-98), the National Natural Science Foundation of China (Project No. 62062058), and the R{\&}D Program of Beijing Municipal Education Commission (Project No. KM202210012002). I would like to thank Shuyang Gu, Shujie Liu, Qinwen Wei and Fang Yan for their valuable assistance with my research experiments and paper writing.



\bibliography{egbib}

\begin{thebibliography}{29}
\providecommand{\natexlab}[1]{#1}
\providecommand{\url}[1]{\texttt{#1}}
\expandafter\ifx\csname urlstyle\endcsname\relax
  \providecommand{\doi}[1]{doi: #1}\else
  \providecommand{\doi}{doi: \begingroup \urlstyle{rm}\Url}\fi

\bibitem[Ak et~al.(2019)Ak, Lim, Tham, and Kassim]{ak2019attribute}
Kenan~E Ak, Joo~Hwee Lim, Jo~Yew Tham, and Ashraf~A Kassim.
\newblock Attribute manipulation generative adversarial networks for fashion
  images.
\newblock In \emph{Proceedings of the IEEE/CVF International Conference on
  Computer Vision}, pages 10541--10550, 2019.

\bibitem[Amir et~al.(2021)Amir, Gandelsman, Bagon, and Dekel]{amir2021deep}
Shir Amir, Yossi Gandelsman, Shai Bagon, and Tali Dekel.
\newblock Deep vit features as dense visual descriptors.
\newblock \emph{arXiv preprint arXiv:2112.05814}, 2\penalty0 (3):\penalty0 4,
  2021.

\bibitem[Avrahami et~al.(2023)Avrahami, Fried, and
  Lischinski]{avrahami2023blended}
Omri Avrahami, Ohad Fried, and Dani Lischinski.
\newblock Blended latent diffusion.
\newblock \emph{ACM Transactions on Graphics (TOG)}, 42\penalty0 (4):\penalty0
  1--11, 2023.

\bibitem[Baldrati et~al.(2023)Baldrati, Morelli, Cartella, Cornia, Bertini, and
  Cucchiara]{baldrati2023multimodal}
Alberto Baldrati, Davide Morelli, Giuseppe Cartella, Marcella Cornia, Marco
  Bertini, and Rita Cucchiara.
\newblock Multimodal garment designer: Human-centric latent diffusion models
  for fashion image editing.
\newblock In \emph{Proceedings of the IEEE/CVF International Conference on
  Computer Vision}, pages 23393--23402, 2023.

\bibitem[Cao et~al.(2023)Cao, Chai, Hao, Zhang, Chen, and
  Wang]{cao2023difffashion}
Shidong Cao, Wenhao Chai, Shengyu Hao, Yanting Zhang, Hangyue Chen, and Gaoang
  Wang.
\newblock Difffashion: Reference-based fashion design with structure-aware
  transfer by diffusion models.
\newblock \emph{IEEE Transactions on Multimedia}, 2023.

\bibitem[Creswell et~al.(2018)Creswell, White, Dumoulin, Arulkumaran, Sengupta,
  and Bharath]{creswell2018generative}
Antonia Creswell, Tom White, Vincent Dumoulin, Kai Arulkumaran, Biswa Sengupta,
  and Anil~A Bharath.
\newblock Generative adversarial networks: An overview.
\newblock \emph{IEEE signal processing magazine}, 35\penalty0 (1):\penalty0
  53--65, 2018.

\bibitem[Ding et~al.(2021)Ding, Yang, Hong, Zheng, Zhou, Yin, Lin, Zou, Shao,
  Yang, et~al.]{ding2021cogview}
Ming Ding, Zhuoyi Yang, Wenyi Hong, Wendi Zheng, Chang Zhou, Da~Yin, Junyang
  Lin, Xu~Zou, Zhou Shao, Hongxia Yang, et~al.
\newblock Cogview: Mastering text-to-image generation via transformers.
\newblock \emph{Advances in Neural Information Processing Systems},
  34:\penalty0 19822--19835, 2021.

\bibitem[Dong et~al.(2020)Dong, Liang, Zhang, Zhang, Shen, Xie, Wu, and
  Yin]{dong2020fashion}
Haoye Dong, Xiaodan Liang, Yixuan Zhang, Xujie Zhang, Xiaohui Shen, Zhenyu Xie,
  Bowen Wu, and Jian Yin.
\newblock Fashion editing with adversarial parsing learning.
\newblock In \emph{Proceedings of the IEEE/CVF conference on computer vision
  and pattern recognition}, pages 8120--8128, 2020.

\bibitem[Feng et~al.(2023)Feng, Zhang, Yu, Fang, Li, Chen, Lu, Liu, Yin, Feng,
  et~al.]{feng2023ernie}
Zhida Feng, Zhenyu Zhang, Xintong Yu, Yewei Fang, Lanxin Li, Xuyi Chen, Yuxiang
  Lu, Jiaxiang Liu, Weichong Yin, Shikun Feng, et~al.
\newblock Ernie-vilg 2.0: Improving text-to-image diffusion model with
  knowledge-enhanced mixture-of-denoising-experts.
\newblock In \emph{Proceedings of the IEEE/CVF Conference on Computer Vision
  and Pattern Recognition}, pages 10135--10145, 2023.

\bibitem[Heusel et~al.(2017)Heusel, Ramsauer, Unterthiner, Nessler, and
  Hochreiter]{heusel2017fid}
Martin Heusel, Hubert Ramsauer, Thomas Unterthiner, Bernhard Nessler, and Sepp
  Hochreiter.
\newblock Gans trained by a two time-scale update rule converge to a local nash
  equilibrium.
\newblock \emph{Advances in neural information processing systems}, 30, 2017.

\bibitem[Ho et~al.(2020)Ho, Jain, and Abbeel]{ho2020denoising}
Jonathan Ho, Ajay Jain, and Pieter Abbeel.
\newblock Denoising diffusion probabilistic models.
\newblock \emph{Advances in neural information processing systems},
  33:\penalty0 6840--6851, 2020.

\bibitem[Hsiao et~al.(2019)Hsiao, Katsman, Wu, Parikh, and
  Grauman]{hsiao2019fashion++}
Wei-Lin Hsiao, Isay Katsman, Chao-Yuan Wu, Devi Parikh, and Kristen Grauman.
\newblock Fashion++: Minimal edits for outfit improvement.
\newblock In \emph{Proceedings of the IEEE/CVF International Conference on
  Computer Vision}, pages 5047--5056, 2019.

\bibitem[Jiang et~al.(2022)Jiang, Yang, Qiu, Wu, Loy, and
  Liu]{jiang2022text2human}
Yuming Jiang, Shuai Yang, Haonan Qiu, Wayne Wu, Chen~Change Loy, and Ziwei Liu.
\newblock Text2human: Text-driven controllable human image generation.
\newblock \emph{ACM Transactions on Graphics (TOG)}, 41\penalty0 (4):\penalty0
  1--11, 2022.

\bibitem[Kang et~al.(2023)Kang, Zhu, Zhang, Park, Shechtman, Paris, and
  Park]{kang2023scaling}
Minguk Kang, Jun-Yan Zhu, Richard Zhang, Jaesik Park, Eli Shechtman, Sylvain
  Paris, and Taesung Park.
\newblock Scaling up gans for text-to-image synthesis.
\newblock In \emph{Proceedings of the IEEE/CVF Conference on Computer Vision
  and Pattern Recognition}, pages 10124--10134, 2023.

\bibitem[Kim et~al.(2023)Kim, Park, Lee, and Choo]{kim2023reference}
Kangyeol Kim, Sunghyun Park, Junsoo Lee, and Jaegul Choo.
\newblock Reference-based image composition with sketch via structure-aware
  diffusion model.
\newblock \emph{arXiv preprint arXiv:2304.09748}, 2023.

\bibitem[Lin et~al.(2023)Lin, Zhao, Ning, Qiu, Wang, and
  Han]{lin2023fashiontex}
Anran Lin, Nanxuan Zhao, Shuliang Ning, Yuda Qiu, Baoyuan Wang, and Xiaoguang
  Han.
\newblock Fashiontex: Controllable virtual try-on with text and texture.
\newblock In \emph{ACM SIGGRAPH 2023 Conference Proceedings}, pages 1--9, 2023.

\bibitem[Men et~al.(2020)Men, Mao, Jiang, Ma, and Lian]{men2020controllable}
Yifang Men, Yiming Mao, Yuning Jiang, Wei-Ying Ma, and Zhouhui Lian.
\newblock Controllable person image synthesis with attribute-decomposed gan.
\newblock In \emph{Proceedings of the IEEE/CVF conference on computer vision
  and pattern recognition}, pages 5084--5093, 2020.

\bibitem[Pernu{\v{s}} et~al.(2023)Pernu{\v{s}}, Fookes, {\v{S}}truc, and
  Dobri{\v{s}}ek]{pernuvs2023fice}
Martin Pernu{\v{s}}, Clinton Fookes, Vitomir {\v{S}}truc, and Simon
  Dobri{\v{s}}ek.
\newblock Fice: Text-conditioned fashion image editing with guided gan
  inversion.
\newblock \emph{arXiv preprint arXiv:2301.02110}, 2023.

\bibitem[Radford et~al.(2021)Radford, Kim, Hallacy, Ramesh, Goh, Agarwal,
  Sastry, Askell, Mishkin, Clark, et~al.]{radford2021clipscore}
Alec Radford, Jong~Wook Kim, Chris Hallacy, Aditya Ramesh, Gabriel Goh,
  Sandhini Agarwal, Girish Sastry, Amanda Askell, Pamela Mishkin, Jack Clark,
  et~al.
\newblock Learning transferable visual models from natural language
  supervision.
\newblock In \emph{International conference on machine learning}, pages
  8748--8763. PMLR, 2021.

\bibitem[Ramesh et~al.(2022)Ramesh, Dhariwal, Nichol, Chu, and
  Chen]{ramesh2022hierarchical}
Aditya Ramesh, Prafulla Dhariwal, Alex Nichol, Casey Chu, and Mark Chen.
\newblock Hierarchical text-conditional image generation with clip latents.
\newblock \emph{arXiv preprint arXiv:2204.06125}, 1\penalty0 (2):\penalty0 3,
  2022.

\bibitem[Rombach et~al.(2022)Rombach, Blattmann, Lorenz, Esser, and
  Ommer]{rombach2022high}
Robin Rombach, Andreas Blattmann, Dominik Lorenz, Patrick Esser, and Bj{\"o}rn
  Ommer.
\newblock High-resolution image synthesis with latent diffusion models.
\newblock In \emph{Proceedings of the IEEE/CVF conference on computer vision
  and pattern recognition}, pages 10684--10695, 2022.

\bibitem[Wu et~al.(2022)Wu, Liang, Ji, Yang, Fang, Jiang, and Duan]{wu2022nuwa}
Chenfei Wu, Jian Liang, Lei Ji, Fan Yang, Yuejian Fang, Daxin Jiang, and Nan
  Duan.
\newblock N{\"u}wa: Visual synthesis pre-training for neural visual world
  creation.
\newblock In \emph{European conference on computer vision}, pages 720--736.
  Springer, 2022.

\bibitem[Yang et~al.(2023{\natexlab{a}})Yang, Gu, Zhang, Zhang, Chen, Sun,
  Chen, and Wen]{yang2023paint}
Binxin Yang, Shuyang Gu, Bo~Zhang, Ting Zhang, Xuejin Chen, Xiaoyan Sun, Dong
  Chen, and Fang Wen.
\newblock Paint by example: Exemplar-based image editing with diffusion models.
\newblock In \emph{Proceedings of the IEEE/CVF Conference on Computer Vision
  and Pattern Recognition}, pages 18381--18391, 2023{\natexlab{a}}.

\bibitem[Yang et~al.(2023{\natexlab{b}})Yang, Chen, and Liao]{unipaint}
Shiyuan Yang, Xiaodong Chen, and Jing Liao.
\newblock Uni-paint: A unified framework for multimodal image inpainting with
  pretrained diffusion model.
\newblock In \emph{Proceedings of the 31st ACM International Conference on
  Multimedia}, MM '23, page 3190–3199, New York, NY, USA, 2023{\natexlab{b}}.
  Association for Computing Machinery.
\newblock ISBN 9798400701085.
\newblock \doi{10.1145/3581783.3612200}.

\bibitem[Zhang et~al.(2021)Zhang, Yin, Fang, Li, Duan, Wu, Sun, Tian, Wu, and
  Wang]{zhang2021ernie}
Han Zhang, Weichong Yin, Yewei Fang, Lanxin Li, Boqiang Duan, Zhihua Wu,
  Yu~Sun, Hao Tian, Hua Wu, and Haifeng Wang.
\newblock Ernie-vilg: Unified generative pre-training for bidirectional
  vision-language generation.
\newblock \emph{arXiv preprint arXiv:2112.15283}, 2021.

\bibitem[Zhang et~al.(2023)Zhang, Rao, and Agrawala]{zhang2023adding}
Lvmin Zhang, Anyi Rao, and Maneesh Agrawala.
\newblock Adding conditional control to text-to-image diffusion models.
\newblock In \emph{Proceedings of the IEEE/CVF International Conference on
  Computer Vision}, pages 3836--3847, 2023.

\bibitem[Zhang et~al.(2018)Zhang, Isola, Efros, Shechtman, and
  Wang]{zhang2018LPIPS}
Richard Zhang, Phillip Isola, Alexei~A Efros, Eli Shechtman, and Oliver Wang.
\newblock The unreasonable effectiveness of deep features as a perceptual
  metric.
\newblock In \emph{Proceedings of the IEEE conference on computer vision and
  pattern recognition}, pages 586--595, 2018.

\bibitem[Zhu et~al.(2017)Zhu, Urtasun, Fidler, Lin, and
  Change~Loy]{zhu2017your}
Shizhan Zhu, Raquel Urtasun, Sanja Fidler, Dahua Lin, and Chen Change~Loy.
\newblock Be your own prada: Fashion synthesis with structural coherence.
\newblock In \emph{Proceedings of the IEEE international conference on computer
  vision}, pages 1680--1688, 2017.

\bibitem[Zhu et~al.(2023)Zhu, Feng, Chen, Bao, Wang, Chen, Yuan, and
  Hua]{zhu2023preerror}
Zixin Zhu, Xuelu Feng, Dongdong Chen, Jianmin Bao, Le~Wang, Yinpeng Chen,
  Lu~Yuan, and Gang Hua.
\newblock Designing a better asymmetric vqgan for stablediffusion, 2023.

\end{thebibliography}
\end{document}